%% file: main.tex
\documentclass[pp]{jmlr}

\input{settings.tex}

\title[Temporal-Clustering Invariance]{Temporal-Clustering Invariance in Irregular Healthcare\\  Time Series}

\author{\Name{Mohammad Taha Bahadori} \Email{bahadorm@amazon.com} 
      \addr Amazon
      \AND
      \Name{Zachary Chase Lipton} \Email{zlipton@cmu.edu} 
      \addr Carnegie Mellon University} 

\vspace{-0.15in}
\begin{document}
\maketitle






\vspace{-0.15in}
\begin{abstract}
\input{abst.tex}
\vspace{-0.1in}
\end{abstract}

\section{Introduction}
\label{sec:intro}
\input{intro.tex}

\vspace{-0.05in}
\section{Methodology}
\label{sec:method}
\input{method.tex}

\section{Experiments}
\label{sec:exp}

\input{exp.tex}

\section{Related Works}
\label{sec:related}
\input{related.tex}


\section{Conclusions and Future Work}
\label{sec:conc}
\input{conc.tex}



\setlength{\bibsep}{1pt}
\small
\bibliography{references}
\bibliographystyle{abbrv}
\normalsize

\clearpage
\newpage
\appendix
\section{Details of the base learners' architecture}
\label{sec:base}
\input{app.tex}


\end{document}

%% file: settings.tex
\usepackage{microtype}
\usepackage{graphicx}
\usepackage{subcaption}
\usepackage{longtable}
\usepackage{subcaption}
\usepackage{booktabs} 

\usepackage{hyperref}

\newcommand{\alg}[2]{\emph{cluster\&count$_{#1}(#2)$}}

\usepackage{bm,amsmath,amssymb}
\usepackage{longtable,lscape}
\usepackage[normalem]{ulem}
\usepackage{booktabs}
\usepackage{xspace}
\usepackage{tcolorbox}
\tcbuselibrary{skins,breakable}
\RequirePackage{color}
\usepackage{array}
\usepackage{sidecap}
\sidecaptionvpos{figure}{c}
\usepackage{natbib}

\usepackage{url}
\usepackage{newfloat}
\usepackage{multirow}
\usepackage{balance}
\usepackage{balance}
\DeclareCaptionType{Algorithm}

\DeclareMathOperator*{\argmin}{argmin}

\usepackage[utf8]{inputenc} 
\usepackage[T1]{fontenc}    
\usepackage{booktabs}       
\usepackage{amsfonts}       
\usepackage{nicefrac}       

\usepackage{algorithm}
\usepackage{algorithmic}
\usepackage{bm}
\usepackage{enumitem}

\newcommand{\xb}{\mathbf{x}}

%% file: abst.tex
Electronic records contain sequences of events,
some of which take place all at once in a single visit,
and others that are dispersed over multiple visits, each with a different timestamp.
We postulate that fine temporal detail,
e.g., whether a series of blood tests 
are completed at once or in rapid succession 
should not alter predictions based on this data. 
Motivated by this intuition, we propose models 
for analyzing sequences of multivariate clinical time series data
that are invariant to this temporal clustering.
We propose an efficient data augmentation technique 
that exploits the postulated temporal-clustering invariance 
to regularize deep neural networks optimized for several clinical prediction tasks. 
We introduce two techniques to temporally coarsen (downsample) irregular time series: 
(i) grouping the data points based on regularly-spaced timestamps; 
and (ii) clustering them, yielding irregularly-paced timestamps. 
Moreover, we propose a MultiResolution Ensemble (MRE) model,
improving predictive accuracy by ensembling predictions 
based on inputs sequences transformed by different coarsening operators.
Our experiments show that MRE improves the mAP 
on the benchmark mortality prediction task from 51.53\% to 53.92\%.



%% file: intro.tex
While common academic sequence learning challenges,
e.g., in natural language processing and speech,
typically consist of evenly-spaced inputs,
in healthcare, 
we frequently encounter multivariate time series, 
where the variables are sampled irregularly
and exhibit signicantly varying base frequencies.
The most popular sequence learning algorithms, 
based on recurrent neural networks, 
act upon discretized time steps, 
with no agreed-upon out-of-the-box method for handling such irregularities.
Thus, to date, deep learning researchers have relied on heuristics,
such as ignoring the timestamp information \citep{choi2016doctor}, 
binning the data into uniformly-spaced time intervals 
\citep{lipton2016modeling, harutyunyan2017multitask}, 
and manual feature extraction \citep{tran2014framework}. 

Several principled approaches have been proposed,
falling broadly into two categories: 
(i) methods that adapt classical methods naturally 
capable of handling irregularly sampled data,
such as point process models \citep{islam2017marked,xiao2017modeling} 
and Gaussian processes 
\citep{li2015classification,shukla2018modeling,futoma2017improved,soleimani2017treatment}---these works extend classical approaches 
with the goal of making their accuracy and scalability 
competitive with modern deep learning approaches;
(ii) methods that adapt recurrent neural networks \citep{che2018recurrent, zheng2017capturing, baytas2017patient,alaa2017learning,ma2017dipole,che2018hierarchical,song2018attend} or convolutional neural networks \citep{razavian2015temporal,nguyen2017mathtt,kooi2017classifying} 
to the irregular time series setting. 

To date, few papers have studied what invariances might hold in irregular time series data,
or mechanisms by which learning algorithms might exploit them.
Suppose a patient visits a hospital for an emergency service 
and undergoes a set of operations. 
Had she visited a smaller clinic, she might have undergone only a subset of those 
and the rest would have been done at a later time in another clinic. 
The first process will result in a single event in the patient's record, 
whereas the second will result in two events. 
Similar patterns can also emerge owing to the timing of the admission, 
insurance policies, or other factors related to the patient's convenience. 
Ideally, a learning algorithm should be robust 
to the temporal clustering of events. 
However, determining the best way to exploit this structure is not straightforward,
and discarding information can be problematic.
For example, the frequency of visits 
is often informative about the severity of illness 
and the algorithms should be able to use it in the analysis. 

As the main contribution of this work, 
we investigate the temporal-clustering properties of clinical time series data. 
To address the existence and usefulness of temporal-clustering invariance, 
we demonstrate data augmentation schemes to exploit it. 
We propose an efficient data augmentation operator 
that coarsens the time series through a stochastic clustering.
The method consists of randomly merging adjacent events,
%
%
merging those more closely-spaced in time with higher probability.
Our experiments demonstrate that this augmentation technique,
in combination with neural network classifiers, 
can yield both more accurate classification 
and greater robustness to variations 
in the temporal-clustering of the data.
The data augmentation pipeline allows us to access 
representations of our inputs sequences
at multiple resolutions throughout training.
We discover that we can boost accuracy further 
by ensembling predictions across input presentation at multiple resolutions. 
Note that this approach requires running 
the multi-resolution pipeline at test time as well.

We explore two different approaches for 
\emph{deterministic} temporal coarsening of irregular time series. 
First, inspired by interpolation networks \citep{futoma2017improved,shukla2018modeling}, 
we shorten the time series to a time series with regularly-spaced timestamps. 
Alternatively, we also propose the \textit{cluster\&count} operator 
to create a shorter version of the irregular time series by clustering the timestamps. 
The values of the new series' data points 
are the averages of the data points belonging to the corresponding clusters. 
We also count the number of data points in each cluster 
and append it to the new time series as an extra feature. 

We can use the two proposed coarsening operators 
to create the MultiResolution Ensemble (MRE) model. 
We select four instances of time series 
using four different coarsening probabilities. 
We process all versions with the same neural network 
and use the attention mechanism to average their predictions. 
The proposed model not only benefits from the event-clustering invariance 
and is more robust to overfitting. 
It shares weights across multiple resolutions of time series 
\citep{che2018hierarchical,razavian2015temporal} 
without a significant increase in the number of parameters. 

Our experiments on two benchmark clinical prediction tasks \citep{harutyunyan2017multitask} 
defined on the publicly available MIMIC-III dataset \citep{johnson2016mimic} 
highlight the usefulness of the proposed operators. 
We test the performance of a large convolutional neural network  with and without data augmentation and show its success 
in improving the generalization of the model. 
Second, we show that data augmentation
improves robustness of the models against adversarial perturbations. 
Finally, our experiments show that MRE improves the mean average percision
on the benchmark mortality prediction task from 51.53\% to 53.92\%.

%% file: method.tex
In this section, we describe the structure 
of sequential EHR data, introduce required notation, 
and then describe our proposed methods.


\paragraph{EHR Structure and our Notation} 
The EHR data for each patient consists of 
a time-labeled multivariate sequence observations. 
Assuming that we use $r$ different variables, 
the time series data for the $n$-th patient $X^{(n)}$ (out of $N$ total patients)
can be represented by a sequence of $T^{(n)}$ tuples 
$(t_i^{(n)}, \mathbf{x}_i^{(n)}) \in \mathbb{R}\times\mathbb{R}^{r} , i=1, \ldots, T^{(n)}$. The timestamp $t_i^{(n)}$ denotes the time 
of the $i$-th visit of the $n$-th patient, 
$\mathbf{x}_i^{(n)}$ denotes an $r$-dimensional feature vector
that typically takes the value of the observed variable(s)
at the corresponding component(s) 
(the treatment of missing variables is addressed in greater detail below),
and $T^{(n)}$ denotes the number of visits of the $n$-th patient.  
To minimize clutter, we drop the superscript $(n)$ whenever it is unambiguous. 
Depending on which precise task we are addressing,
the objective of our predictive models is to predict 
either (i) the label corresponding to each time step  
$\mathbf{y}_i \in \{0,1\}^{s}$ given only the left-ward context, 
or (ii) to predict a sequence-level label at the end of the sequence 
$\mathbf{y} \in \{0,1\}^{s}$. 
Commonly, as in the learning to diagnose task,
the prediction task is multilabel, i.e. $s > 1$.

\paragraph{Learning Tasks} 
Our proposed methods can be used in both learning to diagnose (L2D) \citep{lipton2016learning} 
and encounter sequence modeling (ESM) \citep{choi2016doctor}.
In the L2D task, the input vector $\xb_i$ consists of continuous clinical measurements. 
If there are $r$ different measurements, then $\xb_i \in \mathbb{R}^{r}$. 
The goal of L2D is, given an input sequence $\xb_1, \ldots, \xb_T$, 
to predict the occurrence of a specific disease ($s=1$) or multiple diseases ($s > 1$). 
Without loss of generality, we will describe the algorithm for L2D, 
as ESM can be seen as a special case of L2D 
where we make a prediction at each timestamp $t$ 
given the patient's history before $t$. 

\paragraph{Invariant Learning} Suppose $\mathcal{T}(X)$ denote a transformation of the patient data $X$. If $P(y|\mathcal{T}(X)) = P(y|X)$, prediction of $y$ with $X$ is invariance to transformation $\mathcal{T}$. 
If the relationship holds approximately, i.e. $P(y|\mathcal{T}(X)) \approx P(y|X)$, we have a more relaxed approximate invariance.
Given powerful estimators such as RNNs, we can augment the training data using the  transformation and achieve a $\mathcal{T}-$invariant model. In the rest of this section, we introduce the temporal-clustering approximate invariance, use it in the augmentation setting, and propose an ensemble model based on it.

\begin{algorithm}[tb]
  \caption{Fast Data Augmentation}
  \label{alg:augment}
\begin{algorithmic}[1]
  \STATE \textbf{Input:} patient sequence $\{(t_i, \xb_i, c_i)\}_{i=1}^{T}$,\\ largest clustering probability $p_{\mathrm{high}}$,  Weighted or unweighted augmentation
  \STATE \textbf{Output:} transformed sequence $\{(t'_i, \xb'_i, c'_i)\}_{i=1}^{T'}$.
  \IF{Weighted}
  \STATE $\bm{\Delta} \gets [t_2-t_1, \ldots, t_T-t_{T-1}]$.
  \STATE $\widetilde{\bm{p}}_{\Delta} \gets [1/\Delta_1, \ldots, 1/\Delta_{T-1}]$.
  \STATE $\bm{p}_{\Delta}=\widetilde{\bm{p}}_{\Delta}/(\bm{1}^{\top}\widetilde{\bm{p}}_{\Delta})$.
  \ELSE
  \STATE $\bm{p}_{\Delta} \gets \mathbf{1}_{T-1}/(T-1)$. 
  \ENDIF
  \STATE Draw $p \sim \mathrm{Unif}(0, p_{\mathrm{high}})$
  \STATE  $E\gets \lceil p\cdot T \rceil$ samples without replacement from $\mathrm{Multinomial}(\bm{p}_{\Delta})$
  \STATE Create clusters $\mathcal{C}_{i'}$ for $i'=1, \ldots, T'$ by assigning every pair $(i, i+1)$ to the same cluster if $i \in E$. 
  \FOR{$i'=1$ {\bfseries to} $T'$}
  \STATE $t'_i \gets \mathrm{mean}(\{t: t \in \mathcal{C}_{i'}\})$
  \STATE $\xb'_i \gets \mathrm{mean}(\{\xb: \xb \in \mathcal{C}_{i'}\})$
  \STATE $c'_i \gets \mathrm{sum}(\{c: c \in \mathcal{C}_{i'}\})$
  \ENDFOR
  \STATE \textbf{Return:} sorted sequence $\{(t'_i, \xb'_i, c'_i)\}_{i=1}^{T'}$ based on $t'$.
\end{algorithmic}
\end{algorithm}
\subsection{Fast Data Augmentation}
Data augmentation is the process of creating distorted instances of data points,
nominally increasing the size of the dataset.
When applied to exploit known invariances, 
this technique is well-known to reduce overfitting,
and has become a standard part of the applied computer vision \citep{krizhevsky2012imagenet},
and in automatic speech recognition, 
where the technique is dubbed \emph{multicondition training} \citep{barker2001robust}.
We design a data augmentation procedure for healthcare time series 
demonstrating that temporal clustering is indeed an invariance worth exploiting
in real-world healthcare time series. 

To formally describe the augmentation operator, 
we append a count variable to each tuple in the data, 
i.e. $(t_i, \xb_i) \to (t_i, \xb_i, c_i)$. 
Clearly, $c_i=1$ for $i=1, \ldots, T$ for all original sequences. 
Let $\mathbb{S}_T$ denote the space of all patient sequences of length $T$ 
and a count-augmented patient sequence, 
which is a member of $\mathbb{S}_T$, as $X$. 
The data augmentation operation is a probabilistic mapping $\mathrm{da}_{p}(\cdot): \mathbb{S}_T \mapsto \mathbb{S}_{T'}$, 
where $T' = T - \lceil p\cdot T\rceil$ and $\lceil\cdot\rceil$ is the ceiling operator. 
We draw the coarsening probability $p$ randomly 
according to $p\sim \mathrm{Unif}(0, p_{\mathrm{high}})$,
The upper bound on the transformed sequence length $(1-p_{\mathrm{high}})T$ 
corresponds to the degree of regularization conferred by this approach; 
lower values of $p_{\mathrm{high}}$ give greater regularization.

Given a clustering probability $p$, for a sequence of length $T$, 
we choose $\lceil p\cdot T\rceil$ intervals at random without replacement 
and combine the events at both ends of the selected intervals.  
The details of the algorithm are provided in Algorithm \ref{alg:augment}. 
We further explore a \textit{weighted} version of the data augmentation 
by joining the closer events likelier. 
For the weighted data augmentation, we create a time interval vector of length $T-1$: $\bm{\Delta} = [t_2-t_1, \ldots, t_T-t_{T-1}]$. 
We create a probability value for $T-1$ intervals 
by defining $\widetilde{\bm{p}}_{\Delta} = [1/\Delta_1, \ldots, 1/\Delta_{T-1}]$ 
and normalizing it as $\bm{p}_{\Delta}=\widetilde{\bm{p}}_{\Delta}/(\bm{1}^{\top}\widetilde{\bm{p}}_{\Delta})$. 
We define a multinominal distribution 
using the probability vector $\bm{p}_{\Delta}$ 
and draw $\lceil p\cdot T\rceil$ samples from it without replacement.
We create the final clustering by assigning the data points 
on the ends of each drawn interval to the same cluster. 

The data augmentation combats overfitting 
by providing distorted versions of the data to the training algorithm. 
In typical data augmentation schemes the model 
only uses the unperturbed original data at test time. 
In our augmentation scheme experiments confirm
benefits to providing access to the coarser-grained versions 
of the data at test time. 
To begin, we define our new deterministic time series coarsening operators.

\begin{figure}
\vspace{-0.1in}
    \centering
    \includegraphics[scale=0.25]{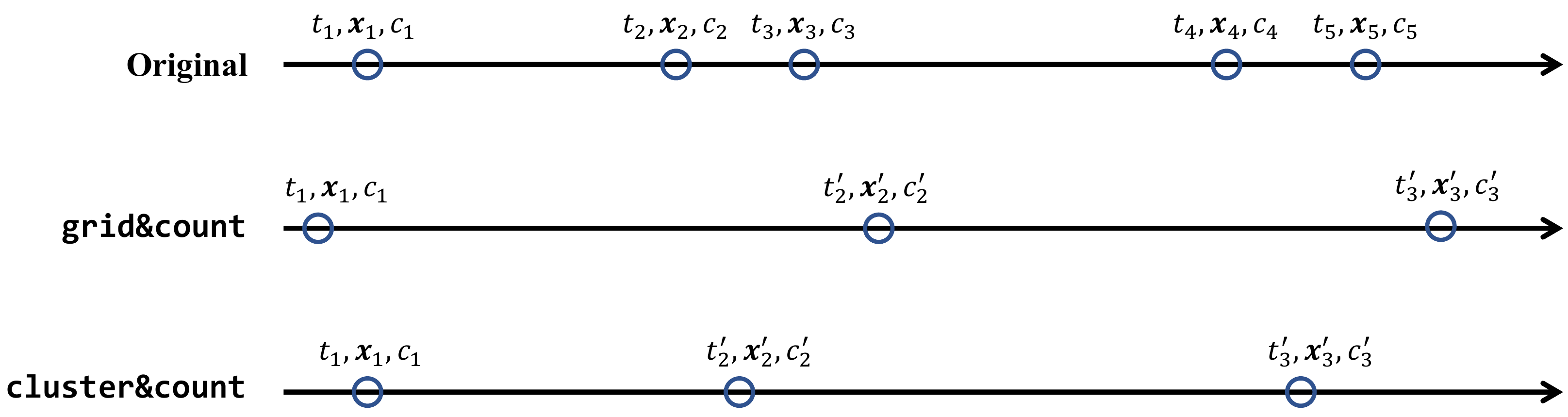}
    \caption{Visualization of the coarsening operators. $grid\&count_{0.6}(\cdot)$ clusters data points into uniformly spaced data points over time. \alg{0.6}{\cdot} produces another irregular time series with the new time stamps $t'_2 = (t_2+t_3)/2$ and $t'_3 = (t_4+t_5)/2$. In both cases we have $\xb'_2 = (\xb_2+\xb_3)/2$ and $\xb'_3 = (\xb_4+\xb_5)/2$. 
    The count variables take values $c'_1 = 1$ and $c'_2 = c'_3 = 2$.
    }
    \label{fig:operator}
\end{figure}

\begin{algorithm}[tb]
  \caption{Grid\&count}
  \label{alg:grid}
\begin{algorithmic}[1]
  \STATE \textbf{Input:} patient sequence $\{(t_i, \xb_i, c_i)\}_{i=1}^{T}$,\\ clustering probability $p$, and observation interval $[t_L, t_R]$.
  \STATE \textbf{Output:} transformed sequence $\{(t'_i, \xb'_i, c'_i)\}_{i=1}^{T'}$.
  \STATE $T' \gets \lceil p\cdot T\rceil$.
  \STATE $t_{i'+1} \gets t_L + i\cdot(t_R-t_L)/(T'-1)$ for $i'=0, \ldots, T'-1$.
  \STATE Initialize clusters $\mathcal{C}_{i'}=\emptyset$ for $i'=1, \ldots, T'$.
  \FOR{$i=1$ {\bfseries to} $T'$}
  \STATE $i^\star \gets \argmin_{i'} \{|t_i-t'_{i'}|\textrm{ for}\, i' = 1, \ldots, T'\}$.
  \STATE $\mathcal{C}_{i^{\star}} \gets \mathcal{C}_{i^{\star}} \cup \{(t_i, \mathbf{x}_i, c_i)\}$
  \ENDFOR
  \FOR{$i'=1$ {\bfseries to} $T'$}
  \STATE $t'_i \gets \mathrm{mean}(\{t: t \in \mathcal{C}_{i'}\})$
  \STATE $\xb'_i \gets \mathrm{mean}(\{\xb: \xb \in \mathcal{C}_{i'}\})$
  \STATE $c'_i \gets \mathrm{sum}(\{c: c \in \mathcal{C}_{i'}\})$
  \ENDFOR
  \STATE \textbf{Return:} sorted sequence $\{(t'_i, \xb'_i, c'_i)\}_{i=1}^{T'}$ based on $t'$.
\end{algorithmic}
\end{algorithm}

\begin{algorithm}[tb]
  \caption{Cluster\&count}
  \label{alg:cluster}
\begin{algorithmic}[1]
  \STATE \textbf{Input:} patient sequence $\{(t_i, \xb_i, c_i)\}_{i=1}^{T}$,\\ clustering probability $p$.
  \STATE \textbf{Output:} transformed sequence $\{(t'_i, \xb'_i, c'_i)\}_{i=1}^{T'}$.
  \STATE $T' \gets \lceil p\cdot T\rceil$.
  \STATE Create clusters $\mathcal{C}_{i'}$ for $i'=1, \ldots, T'$ by clustering $\{t_i\}_{i=1}^{T}$ into $T'$ clusters using $k$-means. 
  \FOR{$i'=1$ {\bfseries to} $T'$}
  \STATE $t'_i \gets \mathrm{mean}(\{t: t \in \mathcal{C}_{i'}\})$
  \STATE $\xb'_i \gets \mathrm{mean}(\{\xb: \xb \in \mathcal{C}_{i'}\})$
  \STATE $c'_i \gets \mathrm{sum}(\{c: c \in \mathcal{C}_{i'}\})$
  \ENDFOR
  \STATE \textbf{Return:} sorted sequence $\{(t'_i, \xb'_i, c'_i)\}_{i=1}^{T'}$ based on $t'$.
\end{algorithmic}
\end{algorithm}

\subsection{The Coarsening Operations}
In designing the coarsening operator, 
we seek transformations from given data sequences
to various different but plausible sequence. 
We exploit our intuition that multiple data points 
might be collected during a single visit or across multiple visits. 
This leads us to the idea of creating different resolutions of the sequence 
by clustering the timestamps into a number of groups,
(resulting in a shorter sequence than the original). 
We note that clustering might significantly change 
the point process properties of the sequence. 
Thus, we append a \emph{count} variable to each cluster 
that indicates the number of data points assigned to that cluster. 
We consider two possibilities for coarsening of irregular time series: 
\emph{regular-grid-based} and \emph{clustering-based}. 

With the grid-based coarsening, 
our aim is to convert an irregular time series 
to a shorter time series with regularly spaced timestamps. 
We first select a set of timestamps, and then assign 
each event to the closest timestamp. 
Next, we compute the average values among events 
assigned to each new timestamp to calculate the transformed time series.
If no event is assigned to a particular timestamp, 
we leave the values as all zeros. 
Finally, we append the cluster size as an extra feature to the feature vector. 
Notice that grid-based coarsening, 
similar to interpolation networks \citep{futoma2017improved,shukla2018modeling}, 
converts irregular time series to regular ones, 
which can make it easier for analysis in the next steps. 
The details of the \textit{grid\&count} operation 
are provided in Algorithm \ref{alg:grid}.

Our proposed alternative is a clustering-based coarsening operator, 
which we denote \emph{cluster\&count}.
Here, we execute a temporal clustering of the sequence of events in the original sequence,
yielding taking the cluster centers as new time stamps. 
The count variables $c_{i'}$ in the transformed sequence 
represent the number of points in the $i'$th cluster. 
Assuming that all categorical features 
have been encoded with one-hot or similar encoding, 
the new feature vectors is the average of all vectors in the same cluster.

Algorithms \ref{alg:grid} and \ref{alg:cluster} 
show the details of the \emph{grid\&count} 
and \emph{cluster\&count} operations, respectively. 
Figure \ref{fig:operator} visualizes the operators in a simple case.
In both coarsening operations, we ignore the missing data during aggregation 
(inside the last for loop in the algorithms). 
If a variable is missed in all events belonging to a cluster, we impute it with zero, following the observations in \citep{lipton2016modeling}.


\subsection{A Multi-resolution Ensemble Network}
In this section, we propose to generate multiple resolutions 
of the input data using the coarsening operators 
described in the previous section for use by our predictive models. 
Our model exploits multi-resolution via an attention mechanism.
To avoid an unnecessary explosion in the number of parameters, 
we share weights of our model across all resolutions.

Formally, let $\bm{f}_{\bm{\theta}}: \mathbb{S}_T \mapsto [0, 1]^s$ 
denote a classifier parameterized by $\bm{\theta}$, 
which maps a patient sequence to probabilities of labels being 1.
Given $K$ different clustering factors $p_k$ for $k=1, \ldots, K$, 
we define the MultiResolution Ensemble (MRE) classifier as follows:
\begin{equation}
\bm{g}_{\bm{\theta}, \bm{\beta}}(X) = \sum_{k=1}^{K} \alpha_{k, \bm{\beta}}(X) \bm{f}_{\bm{\theta}}(C_{p_k}(X)),
\label{eq:mre}
\end{equation}
where $\alpha_{k, \bm{\beta}}(\cdot)$ for $k=1, \ldots, K$ are the attention values. 
The coarsening operator $C_{p_k}(X)$ 
can be either of grid-based or clustering based operators; 
we denote the corresponding networks 
by \textit{MRE-Grid} and \textit{MRE-Cluster}, respectively. 
The attention generation parameters $\bm{\beta}$ 
are the only additional parameters of this classifier. 
We can either use the intermediate hidden representations obtained by $\bm{f_{\theta}}$ 
from the one of $C_{p_k}(X)$ or use simple features such as length of time series. 
Given the pre-determined cluster lengths, 
we can pre-compute the $C_{p_k}(X)$ 
for all sequences and avoid the clustering cost at the run time.

In our experiments, we use MRE with $K=4$ and $p_1 = 1, p_2 = 1/2, p_3 = 1/4, $ and $p_4 = 1/8$. 
The exponentially decreasing $p_k$ is intended to provide access 
to a wide range of resolutions of the input data. 
From an ensemble learning point of view, 
it is beneficial to choose significantly different $p_k$ 
to create more diversity.

Unlike \citet{razavian2015temporal}, who use 
a different CNNs for each resolution of the data, 
MRE shares the weights across all resolutions of the data 
and reduces the number of model parameters. 
We can view MRE as a mixture model \citep{ji2016latent} 
and use the latent-variable inference tools 
to estimate the class probabilities instead of attention variables. 
In this work, we prefer to view the mixture weights as attention weights 
for computational simplicity. 
MRE may potentially be more robust to jitters 
in the timestamps of the sequences \citep{oh2018learning}.

We note a connection between the proposed MRE 
and existing work that studies symmetries and invariances in neural networks \citep{gens2014deep,cohen2016group,kondor2018generalization}.
Similar to the group convolutions, we compute the output of the neural network 
on the input data under different transformations. 
In contrast to the group convolutions, we apply the transformation \textit{grid\&count}  or \textit{cluster\&count} 
to the data rather than the convolutional filters. 
Note that the neither of the coarsening operations 
define a group because it is not invertible 
due to information loss during the clustering operation. 
However, the counting operation attempts to preserve 
more information---related to the point process---during the coarsening process.

\paragraph{Base learners} 
We use the RNN base learners as $\bm{f_{\theta}}$ in Eq. \eqref{eq:mre} 
as they are the common deep neural networks 
for predictive modeling in healthcare time series \citep{lipton2016learning,choi2016doctor}. 
In this architecture, after linear embedding, 
we opt to use two layers of Gated Recurrent Units (GRU) \citep{cho2014learning} 
(instead of the better-known Long Short-Term Memory (LSTM) cells
\citep{hochreiter1997long})
owing to their comparative parsimony with respect to parameters. 
For the prediction layer, we use a residual block with fully connected units. 
We provde further details of the architectures in Appendix \ref{sec:base}.

%% file: exp.tex
We evaluate the proposed algorithms on two L2D  benchmark tasks: 
in-hospital mortality and length of stay (LoS) prediction \citep{harutyunyan2017multitask}, 
which are defined on the publicly available 
MIMIC-III dataset \citep{johnson2016mimic}. 
Both predictive tasks are widely used in 
estimating patients' clinical risk 
and managing costs in hospitals. 
For example, the LACE clinical risk scores  \citep{ben2012simplified,van2012lace+,van2010derivation,gruneir2011unplanned} 
require an estimate of the length of stay of a patient 
soon after they are admitted to the hospital. 
Details of feature extraction and cohort construction 
can be found in the following benchmark paper \citep{harutyunyan2017multitask} 
and we provide a brief summary below.

\subsection{Dataset, tasks, normalization, and training details}

\paragraph{Dataset} 
We follow \citet{harutyunyan2017multitask},
extracting the patient sequences from the MIMIC-III database 
and partition the data into training and testing sets. 
The benchmark data for mortality and LoS prediction tasks have 
17,902/3,236 and 35,344/6,225 training/test data points, respectively. 
We do not use the benchmark discretization and normalization steps 
and normalize the irregular time series by ourselves as follows. 
In the length of stay prediction task, 
we do not use the dataset expansion technique used by the benchmark; i.e.,  
for both training and testing we use 24 hours of patient history.


\paragraph{Normalization of real values} 
Noting the existence of outliers in the data,
we compute the robust mean and variance 
by excluding the values below the 2nd percentile and above 98th percentile. 
Using the estimated robust mean and variance, 
we normalize the real-valued quantities to zero mean and unit variance. 
Additionally, for robustness, we Winsorize the real values 
by thresholding them within the limits of 2nd and 98th percentiles.

\paragraph{Ordinal encoding} 
We use the unary coding \citep{fiete2007neural,moore2011motor} 
for the ordinal quantities such as the Glasgow Coma Score 
instead of the one-hot encoding in \citep{harutyunyan2017multitask}.
A $k$-level ordinal variable with real values $a_1< a_2 < \ldots < a_k$ 
is encoded using a binary vector $\bm{b}$ of size $k-1$. 
We represent each value $a_{\ell}$ for $\ell=1, \ldots, k$ 
by setting the first $\ell-1$ elements of $\bm{b}$ to $1$. 
This approach encodes $a_{\ell+1}-a_{\ell}$ values 
and is able to model large real values for $a_k$. 
We also quantize the count variables $c$ in bins of $(0-1], (1-2], (2, 4], (4, \infty)$ 
and encode them using the unary coding. 
After appending the encoded time and count vectors, 
the resulting feature vectors are $51$-dimensional. 
We use zero-filling to handling missing variables 
because it is simple and efficient \citep{lipton2016modeling}. 

\begin{figure}[!ht]
\vspace{-0.2in}
    \centering
    \subfigure[Mortality, CNN]{\label{fig:cnn_mort}\includegraphics[width=0.45\textwidth]{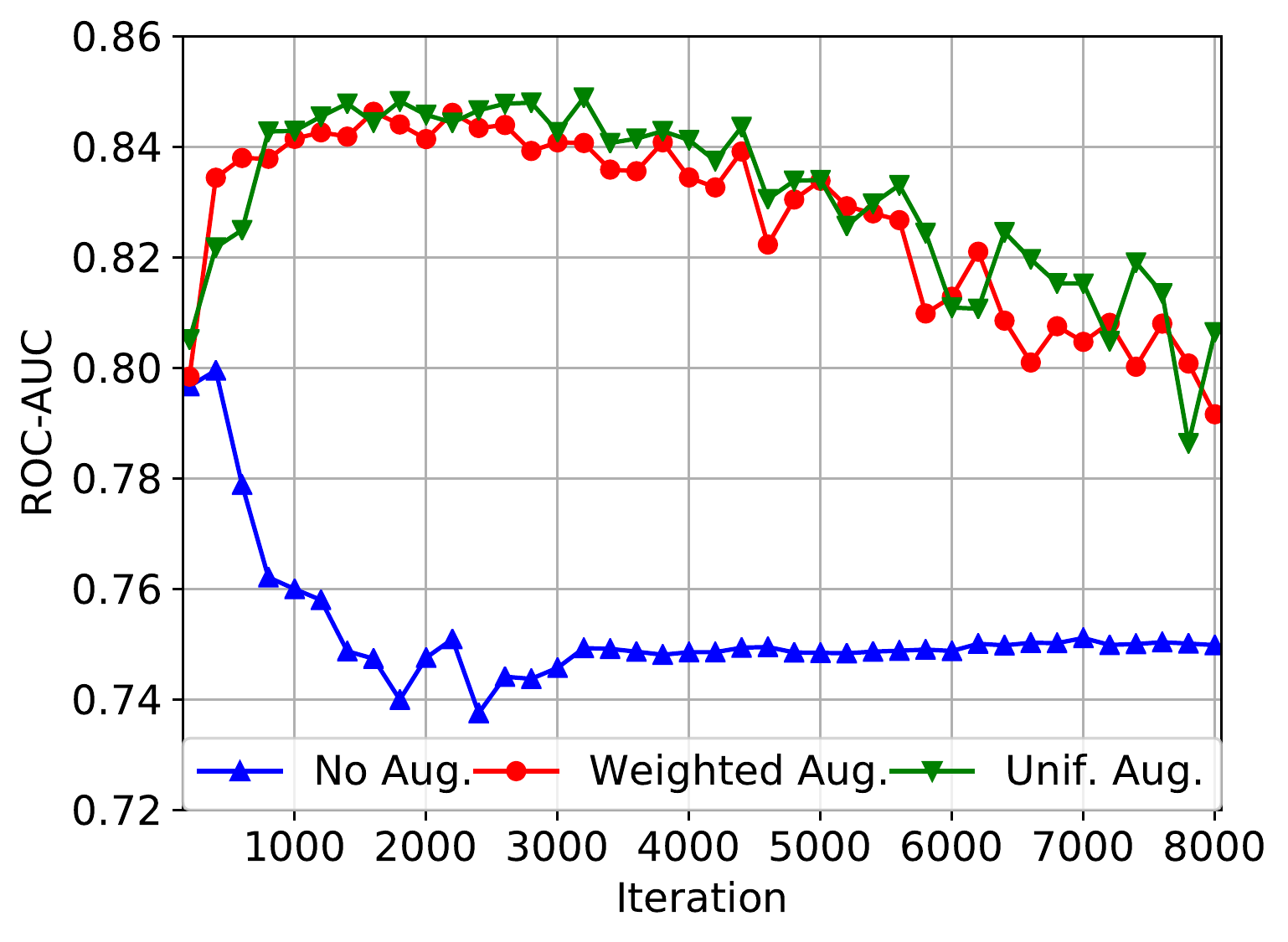}}
    \subfigure[Length of stay, CNN]{\label{fig:cnn_los}\includegraphics[width=0.45\textwidth]{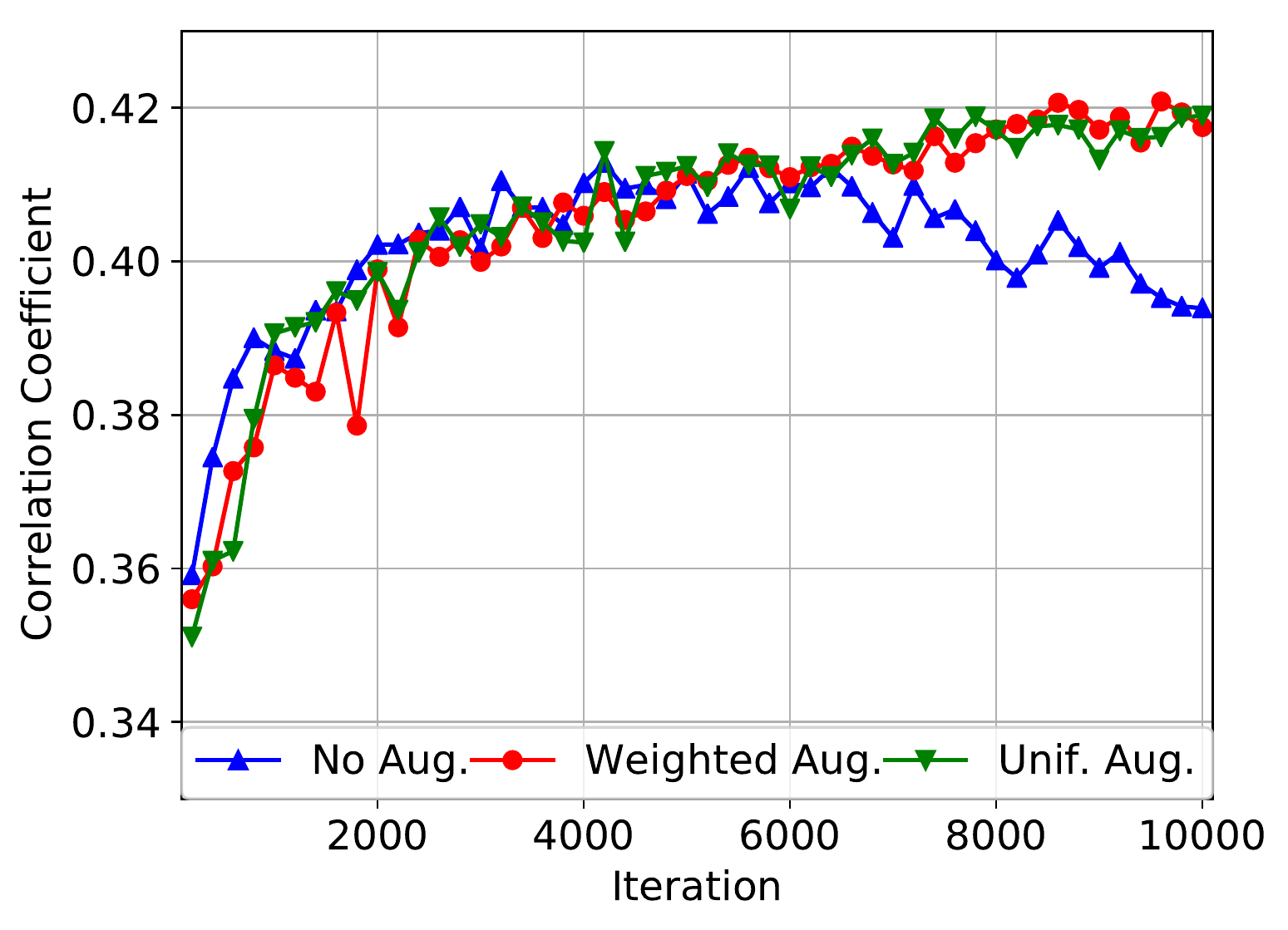}}
    \subfigure[Mortality, RNN]{\label{fig:rnn_mort}\includegraphics[width=0.45\textwidth]{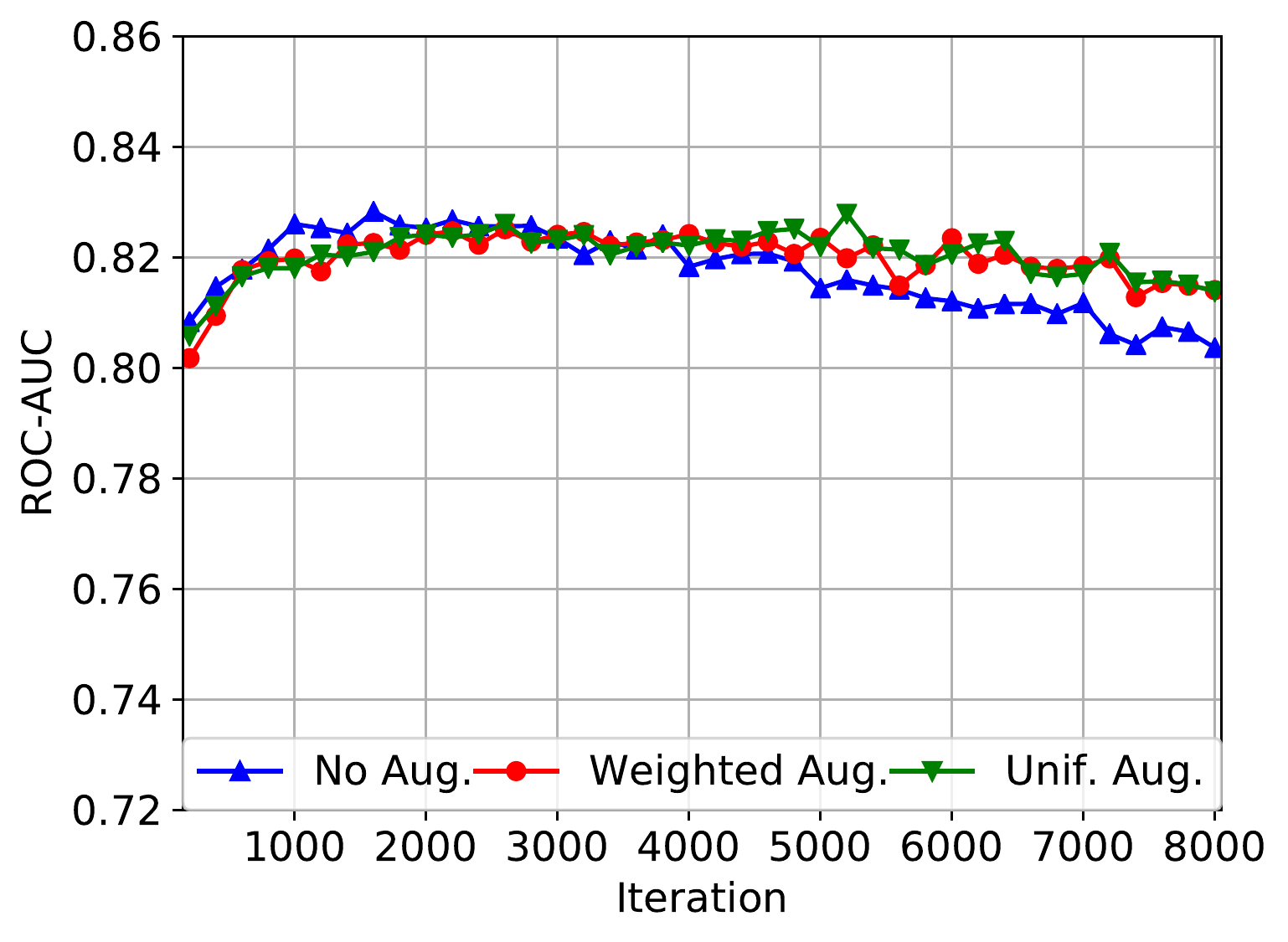}}
    \subfigure[Length of stay, RNN]{\label{fig:rnn_los}\includegraphics[width=0.45\textwidth]{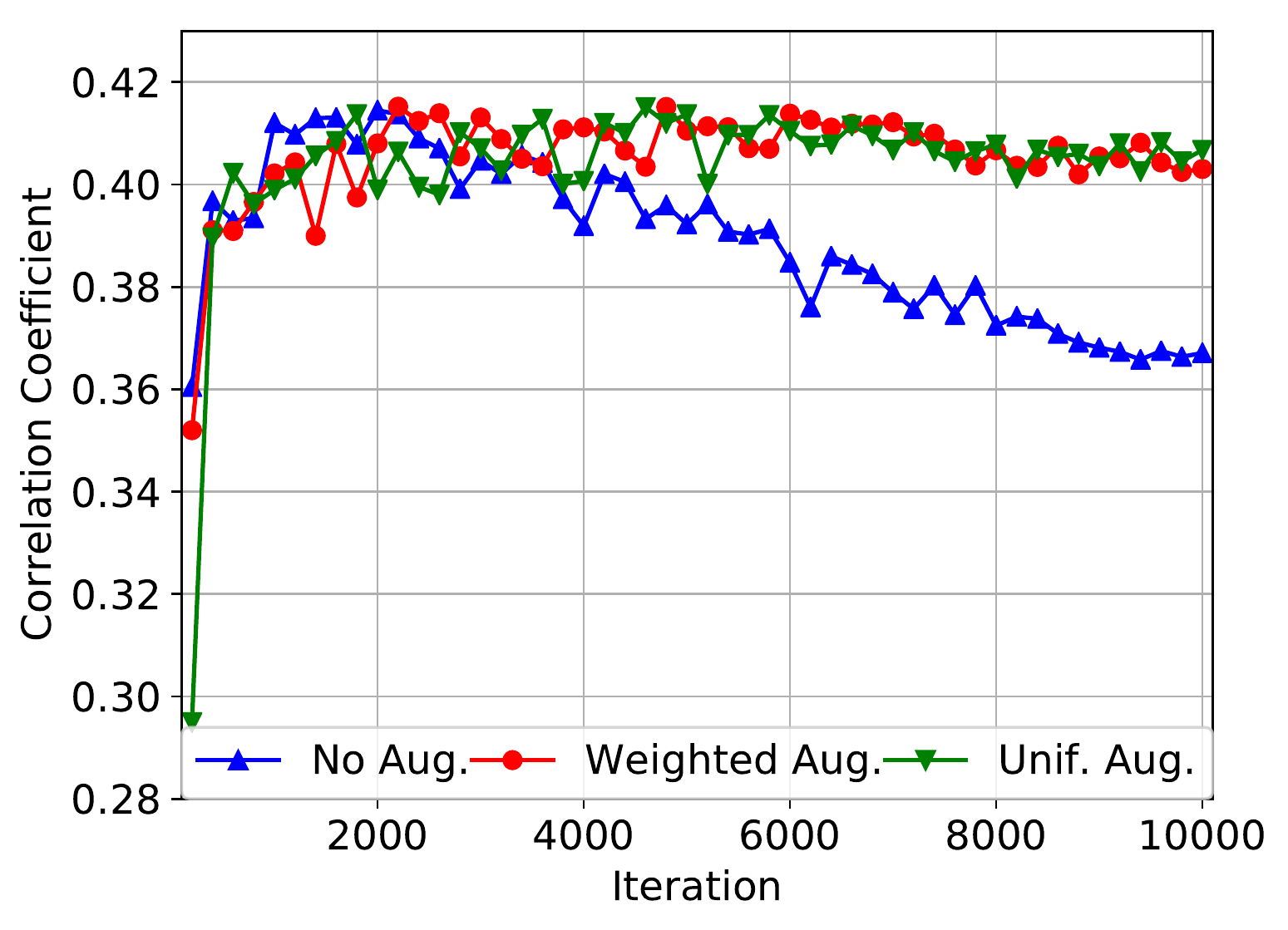}}
    \caption{Augmenting the training data via fast temporal-clustering not only prevents overfitting in large CNN model but also does not significantly hurt the small RNN model. We demonstrate the behavior of the proposed data augmentation on two tasks: mortality and length of stay (LoS) prediction. The CNN model is intentionally chosen to be overparameterized. We choose an overparameterized CNN with and without \emph{cluster\&count} augmentation ($p_{\textrm{high}}=0.5$) and SGD with fixed learning rate and momentum of $10^{-2}$ and $0.95$, respectively. To show smoother curves, we show the average of eight runs with random initializations for each case.}
    \label{fig:augment}
\end{figure}

\paragraph{Training details}
We hold out $15\%$ of the training data as a validation set 
for tuning the hidden layer sizes and hyperparameters 
and report the test results based on the best validation performance.
For optimization, we use Adam \citep{kingma2014adam}
with the AMSGrad modification \citep{reddi2018convergence} with batch size of $100$. 
We halve the learning rate after plateauing for $10$ epochs (determined on validation data)
and stop training after the learning rate drops below $5\times 10^{-6}$.
We run each algorithm $8$ times with different random initializations 
and report the result of the run with the highest validation accuracy.
All of the proposed models are implemented in PyTorch \citep{paszke2017automatic}.

\subsection{Results and analysis}
\paragraph{Data augmentation} 
First, we test the impact of data augmentation 
in improving the robustness of training. 
An ideal data augmentation procedure needs 
to make complex models robust to overfitting, 
without hurting the performance of others. 
To show these desirable attributes for our proposed data augmentation, 
To show the benefits of our methods vis-a-vis overfitting,
we show results on a large CNN that tends generally to overfit this data quickly.
We use the SGD with fixed learning rate and momentum of $10^{-2}$ and $0.95$, respectively. 
Results of this experiments are shown in Figure \ref{fig:augment}. 
To show smoother curves, we show the average of eight runs 
with random initializations for each case. 
Figures \ref{fig:cnn_mort} and \ref{fig:cnn_los} 
show that data augmentation allows longer training time 
before the validation accuracy decreases. 
We also experiment with a GRU model less prone to overfitting. 
Figures \ref{fig:rnn_mort} and \ref{fig:rnn_los} 
show that data augmentation helps the GRU model as well but to a lesser extent. 
More details on both network architectures are given in Appendix \ref{sec:base}.
 
\begin{figure}[ht]
\vspace{-0.2in}
    \centering
    \subfigure[Learning Curve]{\label{fig:learning}\includegraphics[scale=0.47]{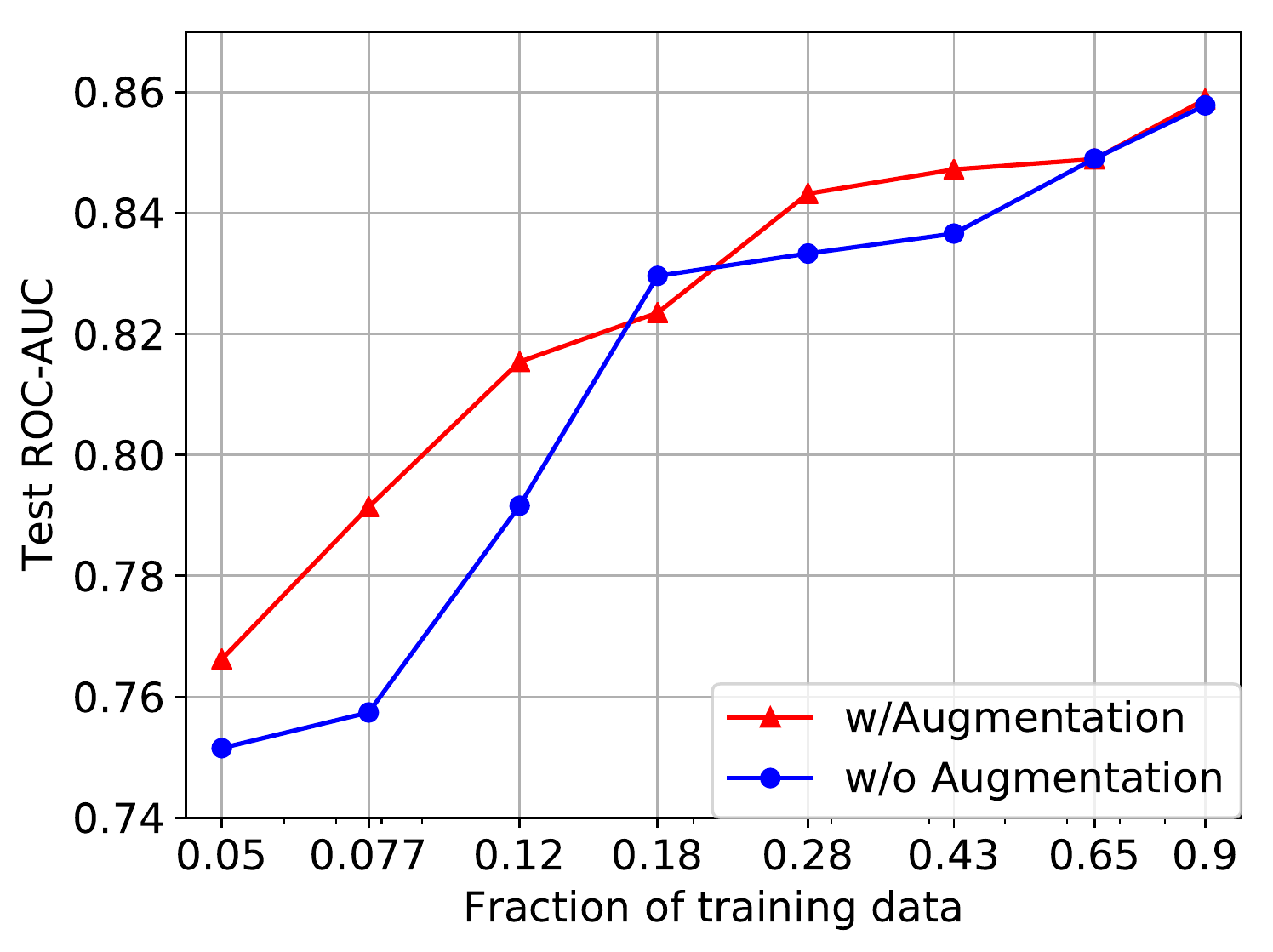}}
    \subfigure[Input Sensitivity Analysis]{\label{fig:adversarial}\includegraphics[scale=0.47]{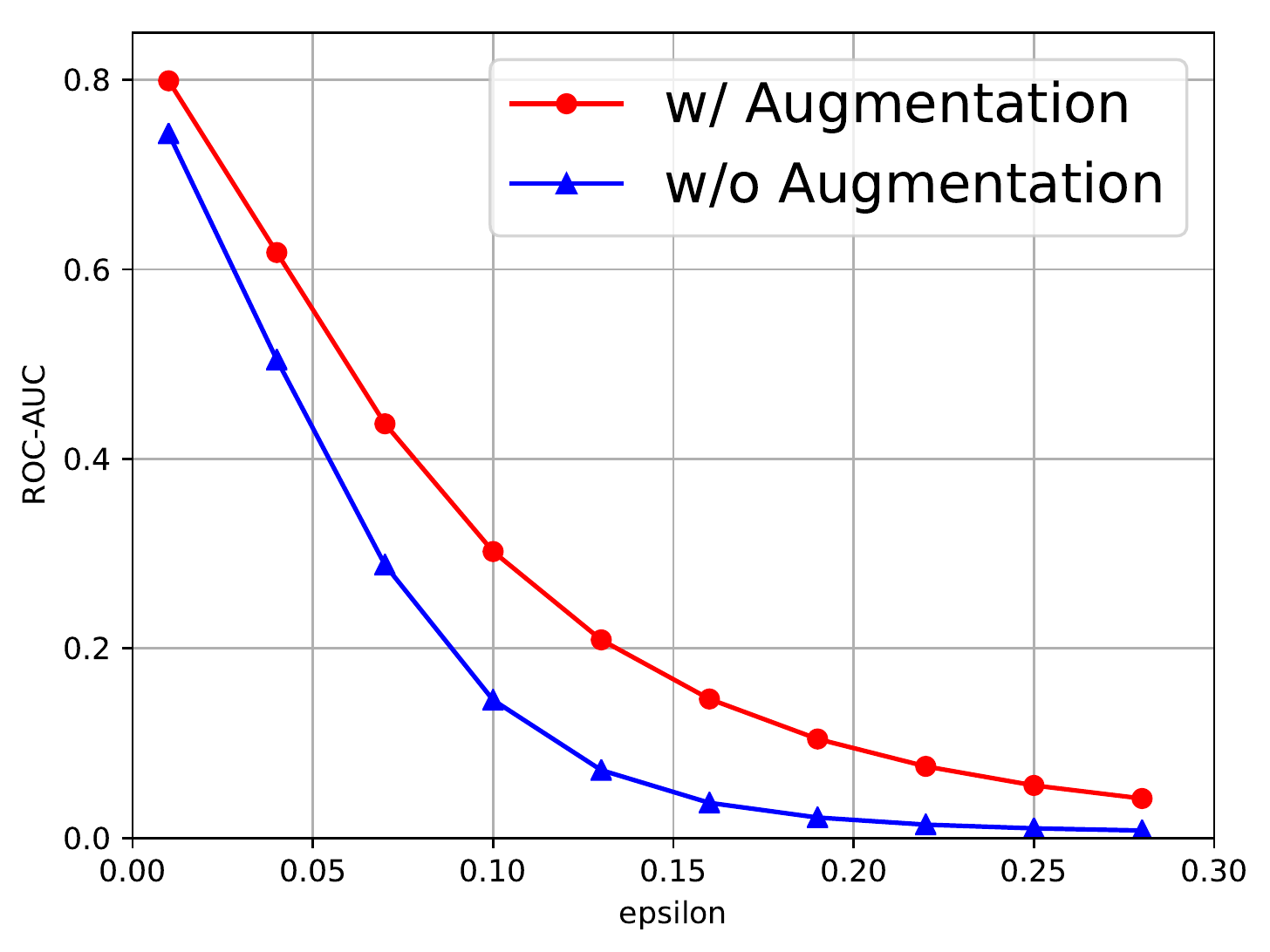}}
    \caption{\textit{(Left)} The learning curve with and without data augmentation on the mortality prediction task. The network is the same GRU network used in Figure \ref{fig:augment}. The x-axis in on a logarithmic scale. \textit{(Right)} Data augmentation using \emph{cluster\&count} improves robustness to adversarial examples. 
    Here we show the ROC-AUC of the CNN network on the mortality task against FGSM attacks 
    of varying strength perturbation $\varepsilon$.}
    \label{fig:two}
\end{figure}

We experiment with both weighted and unweighted coarsening augmentation. 
The results Figure \ref{fig:augment} indicated that 
weighted and unweighted augmentations have similar effects
on the performance of the algorithms. 
Looking at the beginning of Figure \ref{fig:cnn_mort}, 
it is clear that the weighted augmentation 
provides a less diverse set of samples 
that are more similar to the actual data. 
It is a milder data augmentation, 
thus the performance improves faster early during the training. 
Given the milder nature of the weighted augmentation,
it is a better candidate for use alongside other data augmentation techniques.

In Figure \ref{fig:learning}, we examine the impact of data augmentation 
as we increase the fraction of training data used during the training. 
Ignoring the seemingly anomalous data point in $0.18$, 
the general trend indicates that initially when we are given only 5\% of data, 
our data augmentation improves the test AUC by about $1$ percentage point. 
The amount of improvement increases as we use a larger fraction of data 
when we use $0.077$ and $0.12$ fraction of training data points. 
As we further increase the fraction of training data, 
the gain by data augmentation shrinks. 
Note that this this result also indicates that the fact 
that in Figures \ref{fig:rnn_mort} and \ref{fig:rnn_los} 
the gap between non-augmented and augmented training is negligible 
is only because of the size of the training data and the network in the tasks.

Our proposed temporal clustering-based data augmentation 
is more robust to the perturbations on the input features, 
as shown in Figure \ref{fig:adversarial}.
To demonstrate the (relative) adversarial robustness,
we consider its performance under attacks 
via the fast gradient sign method (FGSM) due to \citet{goodfellow2014explaining},
because it is simple and parameterized only with a single parameter $\varepsilon$. 
In this purturbation, we add the adversarial noise $\widetilde{\xb} \gets \xb + \varepsilon\cdot \mathrm{sign}\left( \frac{\partial\mathcal{L}(y, f(\xb))}{\partial \xb}\right)$ to the input, 
where $f(\cdot)$ and $\mathcal{L}(\cdot, \cdot)$ 
denote the prediction and binary cross-entropy loss functions, respectively. 
Then, we evaluate the accuracy of the classifier 
on the perturbed data $f(\widetilde{\xb})$. 
We perform this test on the convolutional neural networks 
in the previous example on the mortality task. 
The results (Figure \ref{fig:adversarial}) 
show that the data augmentation via \emph{cluster\&count} 
does indeed increase the robustness to adversarial perturbations.
The reason behind the robustness is that the time series 
transformed by \emph{cluster\&count} not only have new timestamps 
but also they have new values for the variables.
Note that our analysis is not intended to show 
that our data augmentation creates robustness to adversarial attacks. 
Instead we offer these experiments simply to provide input sensitivity analysis.



\begin{table*}[t]
\caption{In-hospital accuracy results: bold numbers show the best results. Our results are reported in the form of `$\textrm{mean}\,(\textrm{standard error})$'.}
\centering
\begin{tabular}{l|cc}
\toprule
Model & ROC-AUC & mAP\\
\midrule
LSTM \footnotesize\citep{harutyunyan2017multitask}\normalsize & 0.8623 & 0.5153 \\
SAnD \citep{song2018attend} & 0.857 & 0.518 \\
GRU (baseline preprocessing) & 0.8556 (0.000307) & 0.4950 (0.000860) \\
GRU (our preprocessing) & 0.8642 (0.000306) &0.5263 (0.000876) \\
\midrule
GRU-Augmented & 0.8566 (0.000310) & 0.5240 (0.000876) \\
MRE-GRU-Grid & \textbf{0.8670 (0.000310)} & \textbf{0.5392 (0.000879)}\\
MRE-GRU-Cluster & 0.8665 (0.000313) & \textbf{0.5392 (0.000829)}\\
\bottomrule
\end{tabular}
\label{table:mortality_table}
\end{table*}

\begin{table*}[t]
\caption{Length of stay prediction results: bold numbers show the best results. 
Our results are reported in the form of `$\textrm{mean}\,(\textrm{standard error})$'. $\star$-marked experiments use a special dataset expansion technique.}
\setlength{\tabcolsep}{3pt}
\centering
\begin{tabular}{lccc}
\toprule
Model & Correlation & MAE & RMSE \\
\midrule
LSTM \scriptsize\citep{harutyunyan2017multitask}\normalsize  &-- & 94.00$^\star$ & 205.34$^\star$ \\
SAnD \citep{song2018attend}  & -- & -- & 200.93$^\star$\\
GRU (baseline preprocessing) & 0.4093 (0.000478) & 56.3557 (0.0459) & 122.9606 (0.1683) \\
GRU (our preprocessing) & 0.4199 (0.000525)& 55.9366 (0.0504) &123.1219 (0.1760)\\
\midrule
GRU-Augmented & 0.4289 (0.000557) & 55.9823 (0.0514) & 123.1832 (0.1850)\\
MRE-GRU-Grid & 0.4328 (0.000522) & 56.3363 (0.0501) & 123.2220 (0.1627)\\
MRE-GRU-Cluster & \textbf{0.4402 (0.000500)} & \textbf{55.5906 (0.0527)}& 123.6223 (0.1774)\\
\bottomrule
\end{tabular}
\normalsize
\label{table:los_table}
\end{table*}

\paragraph{Accuracy evaluation}
We present the main accuracy results in Tables \ref{table:mortality_table} and \ref{table:los_table}. 
To quantify the randomness in the test set, 
we report the bootstrap ($1000$ runs) estimate of the mean
and standard error of the evaluation measures. 
In the mortality task, we not only include 
the reported single-task result in \citep{harutyunyan2017multitask,song2018attend}, 
but also we report our baseline GRU network on the data 
preprocessed by the benchmark, denoted by ``GRU (base prep)''. 
``GRU (our prep)'' shows the accuracy of our GRU model on the data 
with our special preprocessing describe earlier in this section. 
We measure accuracy of classification with two metrics: 
the area under receiver true-positive vs false-positive curve known as ``ROC-AUC'' 
and the area under precision recall curve 
known as mean average precision or ``mAP''. 
Because the length of stay prediction is a regression task, 
we use three commonly used metrics: mean absolute error (MAE) 
and root mean squared error (RMSE) as un-normalized 
and linear correlation coefficient as a normalized accuracy measures.

As we can see in Tables \ref{table:mortality_table} and \ref{table:los_table}, 
both variants of the proposed MRE together with our preprocessing 
outperforms the single-task baselines 
reported in \citep{harutyunyan2017multitask,song2018attend}. 
Moreover, our GRU network works better than the baseline, 
because of our preprocessing techniques. 
Note that in our ensemble, all weak learners are the same---their 
outputs differ only because they act upon different input representations
We also report the multitask results in the benchmarks only for comparison purposes,
however, comparing them with our single task results may not be fair.

Two observations in the LoS prediction results in Table \ref{table:los_table} stand out: 
(1) Our MAE and RMSE results in the LoS task is much better 
than the results reported in \citep{harutyunyan2017multitask,song2018attend}. 
Inspecting the benchmark preprocessing code, 
we realized that the authors have expanded the training dataset 
by creating examples of patients with time series 
longer than 24 hours and LoS value recalculated at the end of training period. 
We observed that this dataset expansion 
creates significant bias in the results and did not use it; 
(2) Given the large discrepancy between RMSE and MAE results, 
the RMSE values are likely to be impacted by the tail values. 
The standard errors also indicate that  
differences in the results are unlikely to be significant.

\begin{figure}[t]
    \vspace{-0.2in}
     \centering
    \includegraphics[scale=0.47]{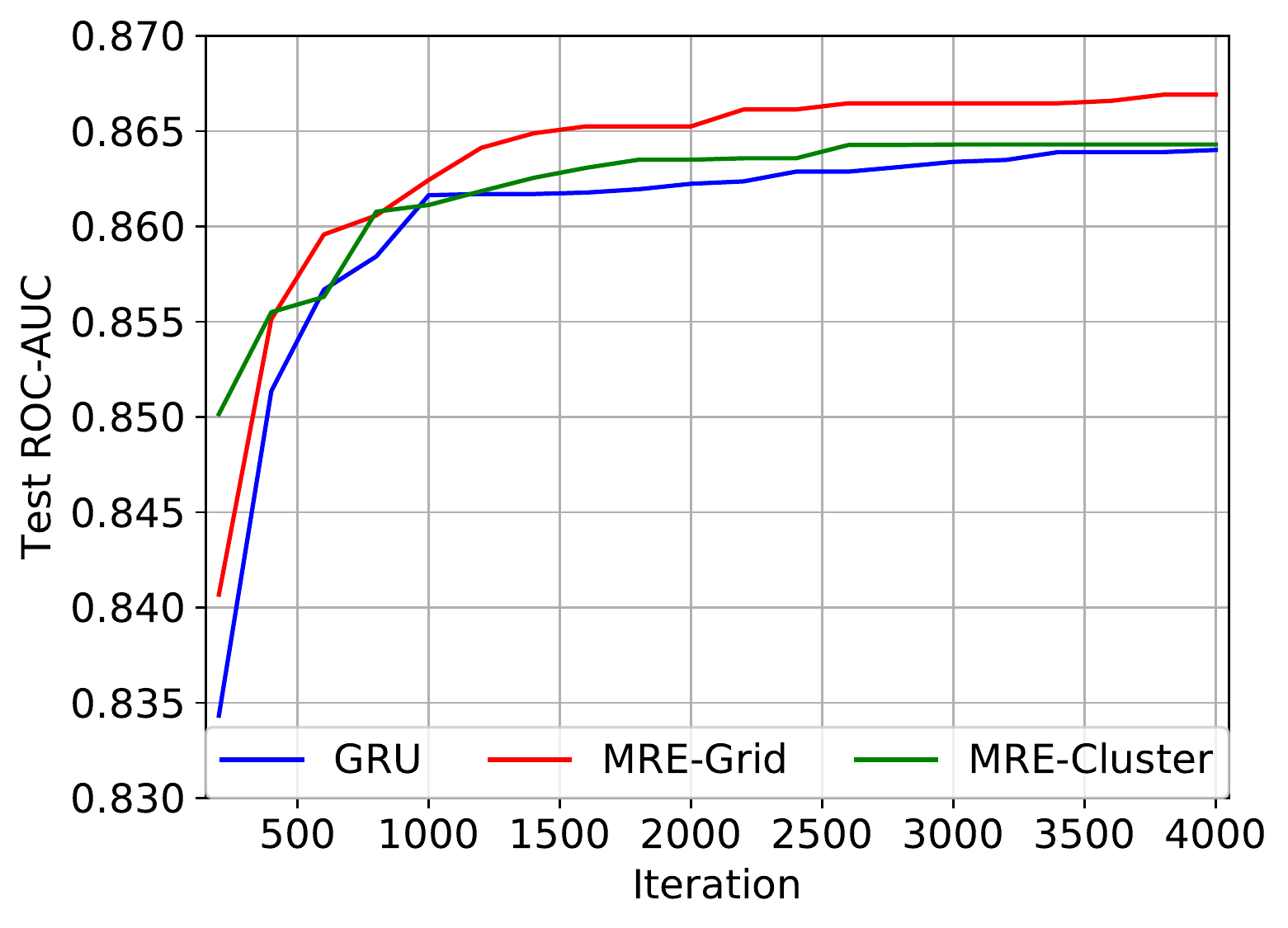}
    \vspace{-0.1in}
    \caption{The MRE-Grid model with the GRU weak-learner 
    converges faster per iteration on the mortality prediction task. 
    This plot shows the average best test accuracy as we train for more iterations. 
    Note that in Tables \ref{table:mortality_table} and \ref{table:los_table} we have used the validation accuracy for selection of the best model, 
    thus the results are different.}
    \label{fig:conv}
    \vspace{-0.15in}
\end{figure}

In terms of speed, MRE is not significantly slower than the corresponding weak learner. 
We run the coarsening operators on the data and persist the clusters 
before the training of MRE and only compute the batches only fly during the training. 
In terms of training convergence speed, 
Figure \ref{fig:conv} shows that MRE converges faster 
compared to the weak learner that it uses.

%% file: related.tex


\vspace{-5px}
\paragraph{Time deformation} 
One well-studied invariance in irregular time series 
is robustness to the deformations in the timing of the events. 
Dynamic Time Warping \citep{vintsyuk1968,sakoe1978dynamic,keogh2005exact}
is a  popular traditional method to incorporate invariance 
to the exact timing of events. 
In more recent work, \citet{oh2018learning} 
propose to learn new timestamps for the events 
to make neural networks invariant to jitters in event times. 
Given the similarity between the coarsening operators 
and the one-dimensional pooling operation, 
we expect the coarsening operation provide a degree of time deformation invariance.

\paragraph{Errors in measurements} 
Measurement errors can happen both due to the devices recording and storing measurements
and due to human error in registering values. 
Adversarial training \citep{goodfellow2014explaining} 
is one effective way to make algorithms less sensitive 
to small errors in the input data. 
If we attach the true timing of the events as a feature to the event features, 
adversarial training can potentially make the algorithm 
robust to jitters in the timing of the events too similar to \citep{oh2018learning}. 
Given that the proposed coarsening operators aggregate the values in the clusters, 
we expect to achieve a degree of invariance to small changes in measurements too.

\paragraph{Missing data} 
Often, not all variables are measured for all patients at all times \citep{lipton2016modeling,che2018recurrent}. 
For data missing completely at random, 
we might seek to make our models robust via dropout   training \citep{srivastava2014dropout},
randomly zeroing out a subset of variables in the input time series 
and augment the data using the newly created data points. Other approaches include inputation via graphical models,
or incorporating missing value indicators
\citep{lipton2016modeling,che2018recurrent}.


\paragraph{Censoring} 
Medical data is often censored, i.e., we do not observe 
the full history for all patients \citep{klein2006survival}. 
For example, in the right censoring scenario, 
if a patient stays alive at the time of a survival analysis task,
her future will not be available to the algorithm. 
Thus, we will not have the full-length records of all patients in our dataset. 
Target replication \citep{lipton2016learning,dai2015semi,ng2015beyond} 
is a technique that requires the algorithm to predict the final outcome 
at any point in the history of a patient. 
Thus, it aims at making the learning algorithm robust to right censoring. 


%% file: conc.tex
In this work, we proposed to exploit a postulated temporal clustering invariance 
and examined the benefits of the derived methods for making predictions given time series of clinical healthcare data. 
Our proposed data augmentation techniques exploit this invariance 
to prevent overfitting in neural networks.
Moreover, we demonstrate that a multi-resolution ensemble technique 
can improve predicted accuracy by acting simultaneously upon multiple resolutions of data.
Our methods in this paper represent clusters by averaging representations.
In future work, we might learn the aggregation function too. 
Additionally, we might seek to understand the relationship 
between temporal clustering invariance and other properties of multivariate clinical time series data.

%% file: app.tex
\paragraph{GRU models}
Our GRU model is a two layers GRU layers followed by a residual block for prediction. The residual block is in the form of relu(bnorm\_2(fc\_2(relu(bnorm\_2(fc\_2($\mathbf{x}$)))))) + fc\_3($\mathbf{x}$). We perform light hyperparameter optimization on the size of hidden vector of GRU.

\paragraph{The CNN model} The CNN model first embeds the input into a higher dimensional vector. Next, it applies a sequence of convolutional residual blocks with progressively larger dilations to the embedded vector. Finally, we flatten the last hidden tensor and apply a batch normalization and dropout layer before  fully connected prediction layer.

In a residual block with dilation factor of $d$, we apply $d$ dilations in the pair of convolutions and $2d$ dilation in the skip connection. The kernel size for all 1d-convs is set to 3.

In the experiments for Figure \ref{fig:augment}, we intentionally large CNN model. Its specific settings are as follows:
\begin{enumerate}
    \item Embedding (dim=200)
    \item ResBlock (200, 180, 160, kernel\_size=3, dil=1)
    \item ResBlock (160, 120, 100, kernel\_size=3, dil=10)
    \item ResBlock (120, 100, 80, kernel\_size=3, dil=25)
    \item FC(1280, 1)
\end{enumerate}

On the length of stay task, we use dilation sequence of (1, 3, 3), because the sequences are shorter.